\documentclass[11pt,letterpaper]{article}
\usepackage[letterpaper]{geometry}
\usepackage{acl2012}
\usepackage{times}
\usepackage[hyphens]{url}
\usepackage{CJK}
\usepackage[draft]{hyperref}
\usepackage{latexsym}
\usepackage{array}
\usepackage{color}

\usepackage{graphicx}
\usepackage[font=normalsize]{subfig}
\usepackage{algorithm}
\usepackage[noend]{algpseudocode}
\usepackage{multirow}

\usepackage{amssymb}
\usepackage{bm}
\usepackage{balance}
\usepackage{arydshln}
\usepackage{mathtools}
\usepackage{soul}

\usepackage{booktabs}
\usepackage{amsmath,amsfonts}
\usepackage{graphicx}
\usepackage{xspace,color}

\usepackage{bbm}

\usepackage{xspace}
\usepackage{textcomp}

\DeclareMathOperator*{\argmax}{arg\,max}

\newcommand{\eg}{\emph{e.g.,}\xspace}
\newcommand{\ie}{\emph{i.e.,}\xspace}
\newcommand{\tp}[1]{#1^\top} 

\title{Learning to Remember Translation History with a Continuous Cache}

\author{Zhaopeng Tu \\ Tencent AI Lab \\ {\normalsize \tt zptu@tencent.com} \And
	      Yang Liu \\ Tsinghua University \\ {\normalsize \tt liuyang2011@tsinghua.edu.cn} \AND
	      Shuming Shi \\ Tencent AI Lab \\ {\normalsize \tt shumingshi@tencent.com} \And
	      Tong Zhang	\\ Tencent AI Lab \\ {\normalsize \tt bradymzhang@tencent.com}}

\begin{document}

\maketitle

\begin{abstract}
Existing neural machine translation (NMT) models generally translate sentences in isolation, missing the opportunity to take advantage of document-level information.
In this work, we propose to augment NMT models with a very light-weight cache-like memory network, which stores recent hidden representations as translation history. The probability distribution over generated words is updated online depending on the translation history retrieved from the memory, endowing NMT models with the capability to dynamically adapt over time.
Experiments on multiple domains with different topics and styles show the effectiveness of the proposed approach with negligible impact on the computational cost.
\end{abstract}

\section{Introduction}

Neural machine translation (NMT) has advanced the state of the art in recent years~\cite{Kalchbrenner:2014:ACL,Cho:2014:EMNLP,Sutskever:2014:NIPS,Bahdanau:2015:ICLR}. However, existing models generally treat documents as a list of independent sentence pairs and ignore cross-sentence information, which leads to translation inconsistency and ambiguity arising from a single source sentence.

There have been few recent attempts to model cross-sentence context for NMT:~\newcite{Wang:2017:EMNLP} use a hierarchical RNN to summarize the previous $K$ source sentences, while~\newcite{Jean:2017:arXiv} use an additional set of an encoder and attention model to dynamically select part of the previous source sentence. While these approaches have proven their ability to represent cross-sentence context, they generate the context from discrete lexicons, thus would cause errors propagated from generated translations. Accordingly, they only take into account source sentences but fail to make use of target-side information.\footnote{\newcite{Wang:2017:EMNLP} indicate that ``considering target-side history inversely harms translation performance, since it suffers from serious error propagation problems.''} Another potential limitation is that they are computationally expensive, which limits the scale of cross-sentence context.


In this work, we propose a very light-weight alternative that can both cover large-scale cross-sentence context as well as exploit bilingual translation history. 
Our work is inspired by recent successes of memory-augmented neural networks on multiple NLP tasks~\cite{Weston:2015:ICLR,Sukhbaatar:2015:NIPS,Miller:2016:EMNLP,Gu:2017:arXiv}, especially the efficient cache-like memory networks for language modeling~\cite{Grave:2017:ICLR,Daniluk:2017:ICLR}.   
Specifically, the proposed approach augments NMT models with {\em a continuous cache} (\textsc{Cache}), which stores recent hidden representations as history context.
By minimizing the computation burden of the cache-like memory, we are able to use larger memory and scale to longer translation history. 
Since we leverage internal representations instead of output words, our approach is more robust to the error propagation problem, and thus can incorporate useful target-side context.

Experimental results show that the proposed approach significantly and consistently improves translation performance over a strong NMT baseline on multiple domains with different topics and styles. 
We found the introduced cache is able to remember translation patterns at different levels of matching and granularity, ranging from exactly matched lexical patterns to fuzzily matched patterns, from word-level patterns to phrase-level patterns.


\section{Neural Machine Translation}
\label{sec-background}

\begin{figure*}[t]
\begin{center}
        \subfloat[Standard NMT]{
            \includegraphics[width=0.21\textwidth]{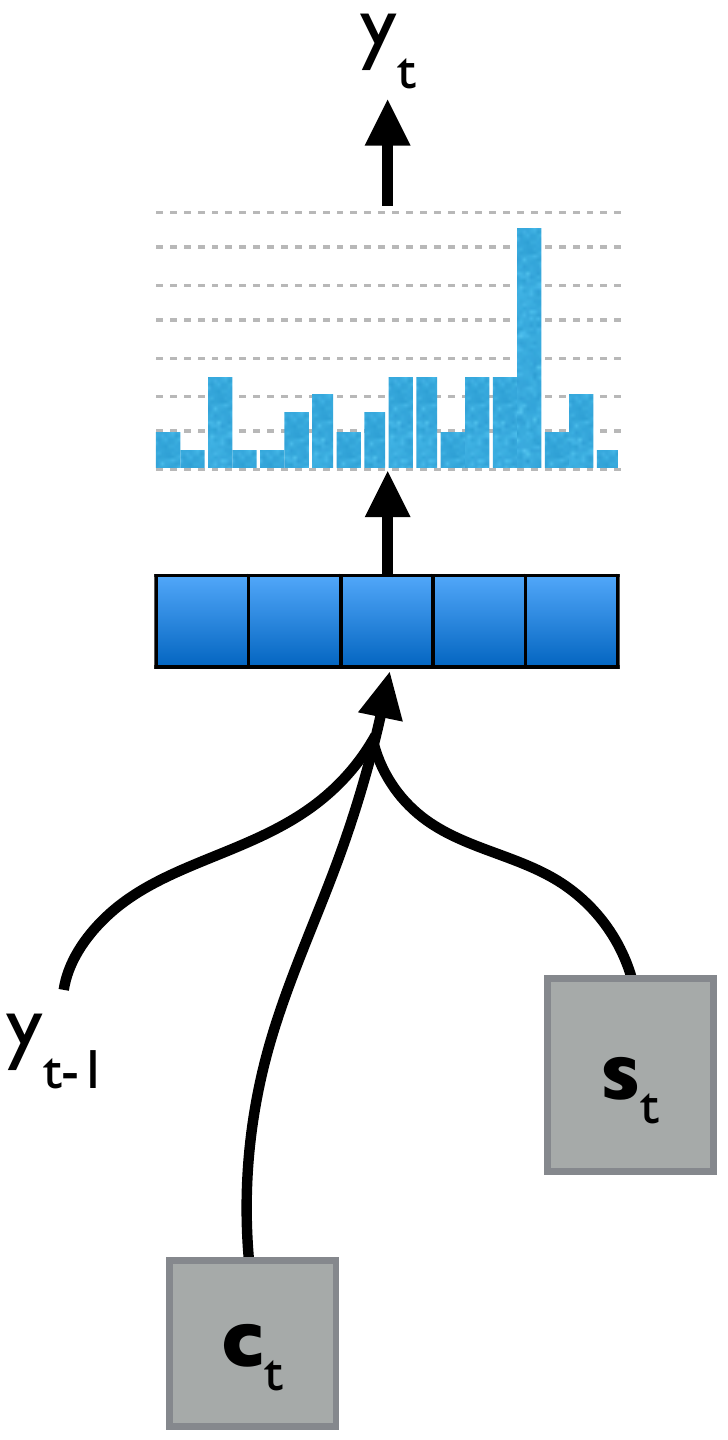}}\hspace{0.2\textwidth}
        \subfloat[NMT augmented with a continuous cache]{
            \includegraphics[width=0.55\textwidth]{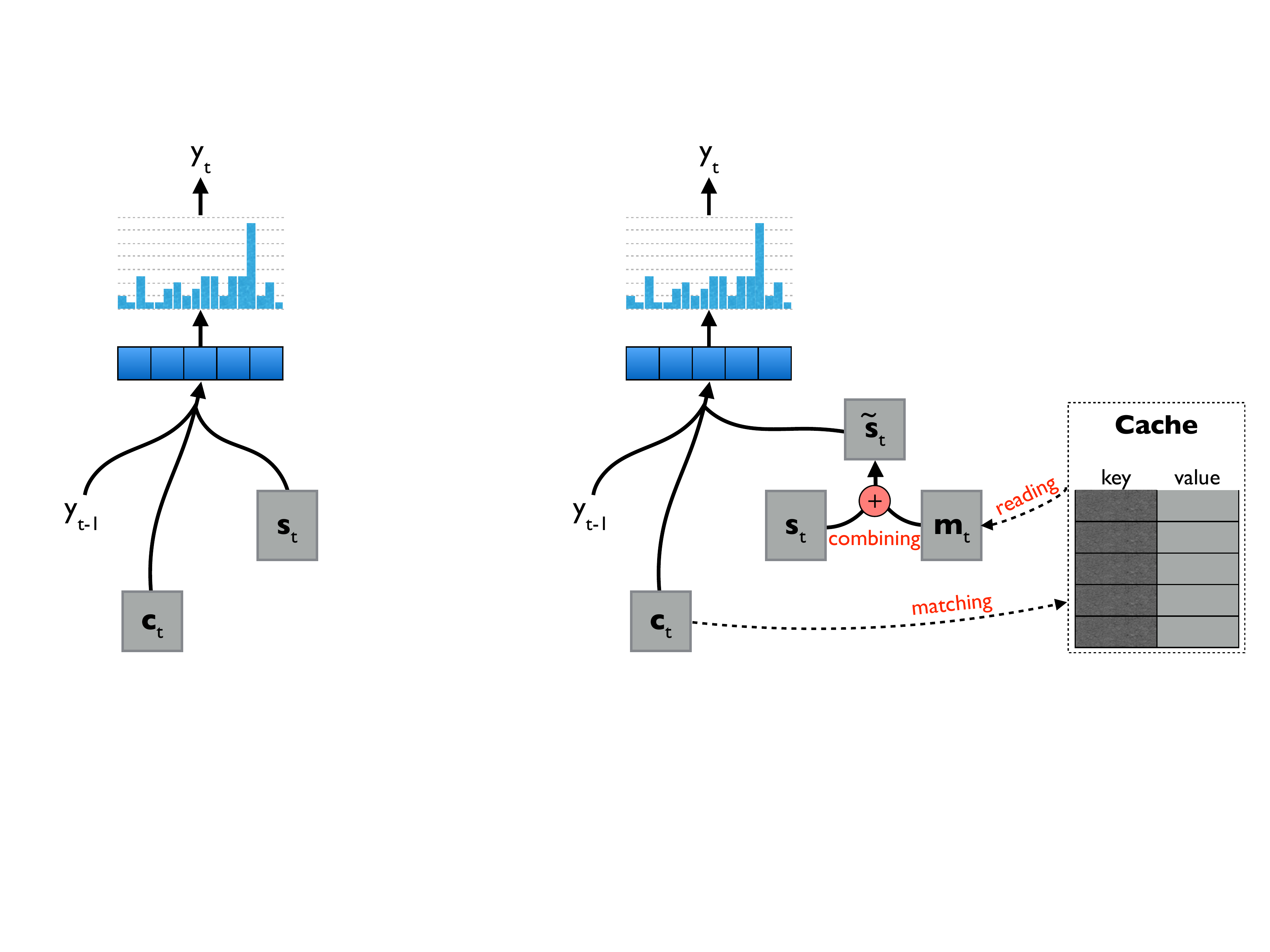}}
\end{center}
\caption{Architectures of (a) standard NMT and (b) NMT augmented with an external cache to exploit translation history. At each decoding step, the current attention context ${\bm c}_t$ that represents source-side content serves as a query to retrieve the cache ({\em key matching}) and an output vector ${\bf  m}_t$ that represents target-side information in the past translations is returned ({\em value reading}), which is combined with the current decoder state ${\bf s}_t$ ({\em representation combining}) to subsequently produce the target word $y_t$.}
\label{figure-architecture}
\end{figure*}

Suppose that ${\bf x}=x_1, \dots x_j, \dots x_J$ represents a source sentence and ${\bf y}=y_1, \dots y_t, \dots y_{T}$ a target sentence. NMT directly models the probability of translation from the source sentence to the target sentence word by word:
\begin{equation}
P({\bf y}|{\bf x}) = \prod_{t=1}^{T} P(y_t| y_{<t}, {\bf x})
\end{equation}
As shown in Figure~\ref{figure-architecture} (a), the probability of generating the {\em t}-th word $y_t$ is computed by
\begin{equation}
P(y_t| y_{<t}, {\bf x}) = g(y_{t-1}, {\bf s}_t, {\bf c}_t)
\label{eqn-standard-probability}
\end{equation}
where $g(\cdot)$ first linearly transforms its input and then applies a softmax function, $y_{t-1}$ is the previously generated word, ${\bf s}_t$ is the $t$-th decoding hidden state, and ${\bf c}_t$ is the $t$-th source representation.
The decoder state ${\bf s}_t$ is computed as follows:
\begin{equation}
{\bf s}_t = f(y_{t-1}, {\bf s}_{t-1}, {\bf c}_t)
\label{eqn-hidden-state}
\end{equation}
where $f(\cdot)$ is an activation function, which is implemented as GRU~\cite{Cho:2014:EMNLP} in this work.
${\bf c}_t$ is a dynamic vector that selectively summarizes certain parts of the source sentence at each decoding step:
\begin{equation}
{\bf c}_t = \sum_{j=1}^{J} \alpha_{t,j} {\bf h}_j
\end{equation}
where $\alpha_{t,j}$ is alignment probability calculated by an attention model~\cite{Bahdanau:2015:ICLR,Luong:2015:EMNLP}, and ${\bf h}_j$ is the encoder hidden state of the $j$-th source word $x_j$.

\begin{CJK}{UTF8}{gbsn}
\begin{figure}[t]
\centering
\subfloat[The translation of ``机遇'' (``{\em opportunity}'') suffers from ambiguity problem, while the translation of ``觉得'' (``{\em feel}'') suffers from tense inconsistency problem. The former problem is {\bf \em not} caused by attending to wrong source words, as shown below.]{
\begin{tabular}{|c|m{6.1cm}|}
\hline
Src & \dots 开始 都 {\bf 觉得} \dots 大家 {\bf 觉得} 这 也是 一次 {\bf 机遇} ， 一次 挑战 。\\\hdashline
Ref & \dots initially they all {\bf felt} that \dots  everyone {\bf felt} that this was also an {\bf opportunity} and a challenge .\\\hdashline
NMT & \dots {\bf \color{blue} felt} that \dots we {\bf  \color{red} feel} that it is also a {\bf \color{red} challenge} and a challenge .\\
\hline
\end{tabular}
} \\
\subfloat[Attention matrix.]{
\includegraphics[width=0.36\textwidth]{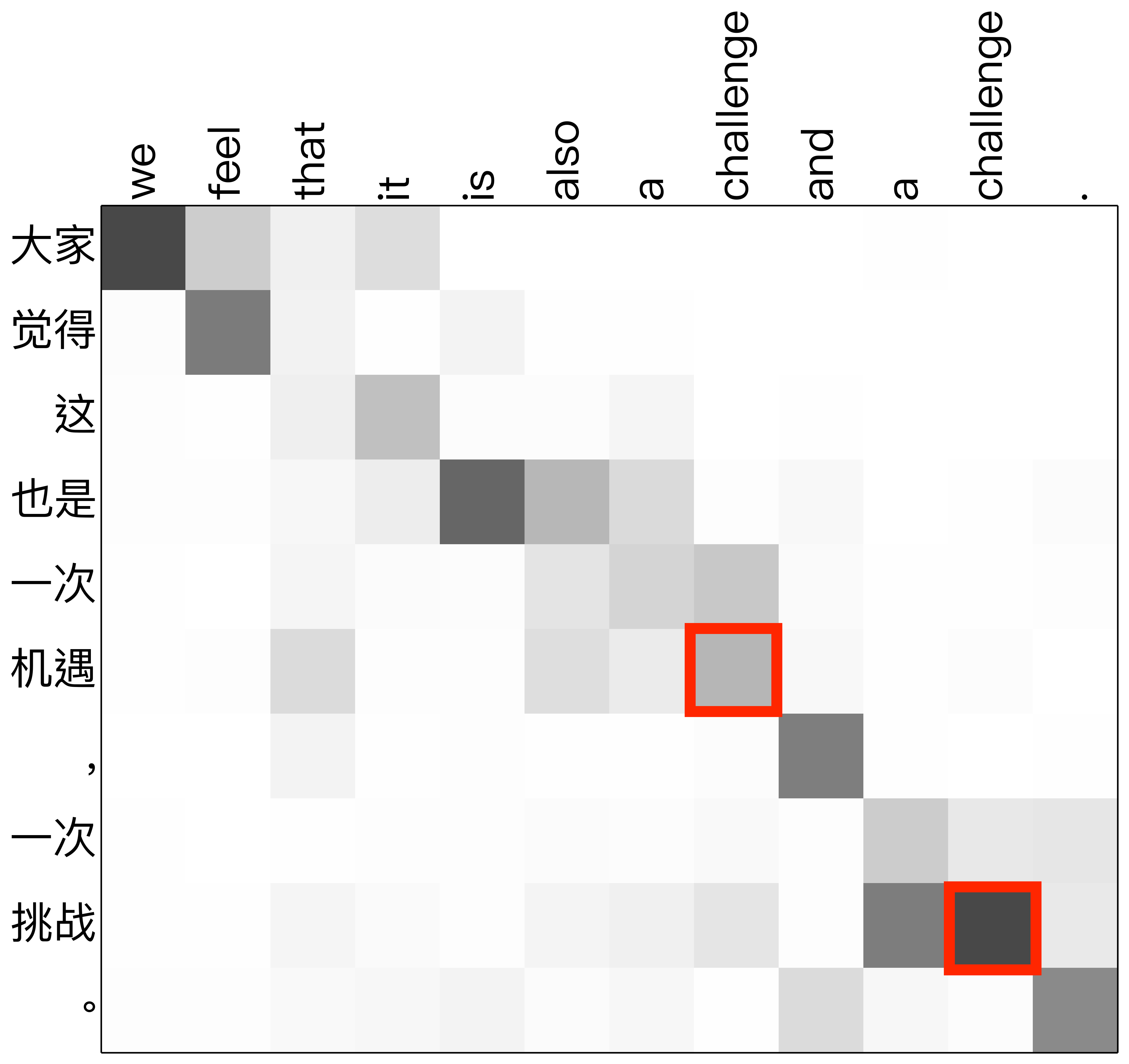}}
\caption{An example translation.}
\label{table-problems}
\end{figure}
\end{CJK}

Since the continuous representation of a symbol (\eg ${\bf h}_j$ and ${\bf s}_t$) encodes multiple meanings of a word, NMT models need to spend a substantial amount of their capacity in disambiguating source and target words based on the context defined by a source sentence~\cite{choi2016context}. 
Consistency is another critical issue in document-level translation, where a repeated term should keep the same translation throughout the whole document \cite{Xiao:2011:MTSummit}.
Nevertheless, current NMT models still process a document by translating each sentence alone, suffering from inconsistency and ambiguity arising from a single source sentence, as shown in Table~\ref{table-problems}. These problems can be alleviated by the proposed approach via modeling translation history, as described below.

\section{Approach}

\subsection{Architecture}


The proposed approach augments neural machine translation models with a cache-like memory, which has proven useful for capturing longer history for the language modeling task~\cite{Grave:2017:ICLR,Daniluk:2017:ICLR}.
The cache-like memory is essentially a key-value memory~\cite{Miller:2016:EMNLP}, which is an array of slots in the form of ({\em key}, {\em value}) pairs. The matching stage is based on the key records while the reading stage uses the value records. 
From here on, we use {\em cache} to denote the cache-like memory.

Since modern NMT models generate translation in a word-by-word manner, translation information is generally stored at word level, including source-side context that embeds content being translated and target-side context that corresponds to the generated word. 
With the goal of remembering translation history in mind, the key should be designed with features to help match it to the source-side context, while the value should be designed with features to help match it to the target-side context.
To this end, we define the cache slots as pairs of vectors $\{({\bf c}_1, {\bf s}_1), \dots, ({\bf c}_i, {\bf s}_i), \dots, ({\bf c}_I, {\bf s}_I)\}$ where ${\bf c}_i$ and ${\bf s}_i$ are the attention context vector and its corresponding decoder state at time step $i$ from the previous translations. The two types of representation vectors correspond well to the source- and target-side contexts~\cite{Tu:2017:TACL}.

Figure~\ref{figure-architecture}(b) illustrates the model architecture. At each decoding step $t$, the current attention context ${\bf c}_t$ serves as a query, which is used to {\em match} and {\em read} from the cache looking for relevant information to generate the target word. The retrieved vector ${\bf m}_t$, which embeds target-side contexts of generating similar words in the translation history, is {\em combined} with the current decoder state ${\bf s}_t$ to subsequently produce the target word $y_t$ (Section 3.2). When the full translation is generated, the decoding contexts are stored in the cache as a history for future translations (Section 3.3).

\subsection{Reading from Cache}

Cache reading involves the following three steps:

\paragraph{Key Matching}
The goal of key matching is to retrieve similar records in the cache. To this end, we exploit the attention context representations ${\bf c}_t$ to define a probability distribution over the records in the cache.
Using context representations as keys in the cache, the cache lookup operator can be implemented with simple dot products between the stored representations and the current one:
\begin{equation}
P_{m}( {\bf c}_i \ | {\bf c}_t) = \frac{\exp (\tp{{\bf c}_t} {\bf c}_i)}{\sum_{i'=1}^{I} \exp (\tp{{\bf c}_t} {\bf c}_i')}
\label{eqn-key-matching}
\end{equation}
where ${\bf c}_t$ is the attention context representation at the current step $t$, ${\bf c}_i$ is the stored representation at the $i$-th slot of the cache, and $I$ is the number of slots in the cache.
In contrast to existing memory-augmented neural networks, the proposed cache avoids the need to learn the memory matching parameters, such as those related to parametric attention models~\cite{Sukhbaatar:2015:NIPS,Daniluk:2017:ICLR}, transformations between the query and keys~\cite{Miller:2016:EMNLP,Gu:2017:arXiv}, or human-defined scalars to control the flatness of the distribution~\cite{Grave:2017:ICLR}.\footnote{We tried these matching implementations in our preliminary experiments, but found no improvements for this task.}

\paragraph{Value Reading}
The values of the cache is read by taking a sum over the stored values ${\bf s}_i$, weighted by the matching probabilities from the keys, and the retrieved vector ${\bf m}_t$ is returned:
\begin{equation*}
{\bf m}_t  = \sum_{({\bf c}_i, {\bf s}_i) \in cache} P_{m}({\bf c}_i \ | {\bf c}_t) ~ {\bf s}_i
\end{equation*}
From the view of memory-augmented neural networks, the matching probability $P_{m}({\bf c}_i \ | {\bf c}_t)$ can be interpreted as the probability to retrieve similar target-side information ${\bf m}_t$ from the cache given the source-side context ${\bf c}_t$, where the desired answer is the contexts related to similar target words generated in past translations.


\paragraph{Representation Combining}
The final decoder state that is used to generate the next-word distribution is computed from a linear combination of the original decoder state ${\bf s}_t$ and the output vector ${\bf m}_t$ retrieved from the cache:\footnote{We tried the strategy of ``Gating Auxiliary Context'' used in~\cite{Wang:2017:EMNLP} in our preliminary experiments, and found similar performance.}
\begin{eqnarray}
& \tilde{\bf s}_t = (\mathbf{1} - {\bm \lambda}_t) \otimes {\bf s}_t + {\bm \lambda}_t \otimes {\bf m}_t \\
& P(y_t| y_{<t}, {\bf x}) = g(y_{t-1}, {\bf c}_t, \tilde{\bf s}_t)
\label{eqn-our-probability}
\end{eqnarray}
where $\otimes$ is an element-wise multiplication, and ${\bm \lambda}_t \in \mathbb{R}^{d}$ is a dynamic weight vector calculated at each decoding step.
This strategy is inspired by the concept of {\em update gate} from GRU~\cite{Cho:2014:EMNLP}, which takes a linear sum between the previous hidden state and the candidate new hidden state.
The starting point for this strategy is an observation: generating target words at different steps has the different needs of the translation history. For example, translation history representation is more useful if a similar slot is retrieved in the cache, while less by other cases. To this end, we calculate the dynamic weight vector by
\begin{equation}
{\bm \lambda}_t =  \sigma ({\bf U} {\bf s}_t + {\bf V} {\bf c}_t + {\bf W} {\bf m}_t)
\label{eqn-context-gate}
\end{equation}
Here $\sigma(\cdot)$ is a logistic sigmoid function, and $\{{\bf U} \in \mathbb{R}^{d\times d}, {\bf V}  \in \mathbb{R}^{d\times l}, {\bf W}  \in \mathbb{R}^{d\times d}\}$ are the new introduced parameter matrices with $d$ and $l$ being the number of units of decoder state and attention context vector, respectively.
Note that ${\bm \lambda}_t$ has the same dimensionality as ${\bf s}_t$ and ${\bf m}_t$, and thus each element in the two vectors has a distinct interpolation weight. In this way, we offer a more precise control to combine the representations, since different elements retain different information.

The addition of the continuous cache to a NMT model inherits the advantages of cache-like memories: the probability distribution over generated words is updated online depending on the translation history, and consistent translations can be generated when they have been seen in the history.
The neural cache also inherits the ability of the decoder hidden states to model longer-term cross-sentence contexts than intra-sentence context, and thus allows for a finer modeling of the document-level context.

\subsection{Writing to Cache}

The cache component is an external key-value memory structure which stores $I$ elements of recent histories, where the key at position $i\in[1,M]$ is ${\bf k}_i$ and its value is ${\bf v}_i$. For each key-value pair, we also store the corresponding target word $y_t$ as an indicator for the following updating operator.
\footnote{In the writing phrase, the cache component works like a standard cache, in which the target word $y_t$ serves as the ``key'' to address the ``value'' (${\bf k}_t$, ${\bf v}_t$) for updating the cache.}

In this work, we focus on learning to remember and exploit cross-sentence translation history. 
Accordingly, different from~\cite{Grave:2017:ICLR,Kawakami:2017:ACL} where the cache is updated after each generation of target word, we write to the cache after a translation sentence is fully generated.
Given a generated translation sentence ${\bf y}=\{y_1,\dots, y_t, \dots, y_T\}$, its corresponding attention vector sequence is $\{{\bf c}_1,\dots, {\bf c}_t,\dots,{\bf c}_T\}$ and the decoder state sequence is $\{{\bf s}_1,\dots, {\bf s}_t,\dots,{\bf s}_T\}$.
Each triple $\langle{\bf c}_t, {\bf s}_t, y_t\rangle$ is written to the cache as follows:
\begin{enumerate}
 \item {\em If $y_t$ does not exist in the cache}, an empty slot is chosen or the least recently used slot is overwritten, where the key slot is ${c}_t$, the value slot is ${\bf s}_t$ and the indicator is $y_t$.
 \item {\em If $y_t$ already exists in the cache at some position $i$}, the key and value are updated: ${\bf k}_i = ({\bf k}_i + {\bf c}_t)/2$ and ${\bf v}_i = ({\bf v}_i + {\bf s}_t)/2$.
\end{enumerate}
From the perspective of ``general cache policy'', it can be regarded as a sort of exponential decay, since at each update the previous keys and values are halved. From the perspective of continuous cache, on the other hand, the intuition behind is to model temporal order for the same word -- the more recent histories serve as more important roles.

Some researchers may worry about that the key ${\bf k}_i$ and the attention vector ${\bf c}_t$ could be fully unrelated, since they ``align'' the same word $y_t$ to the source words of different source sentences. 
We believe that such case would rarely happen.
When ${\bf c}_t$ is aligned to a target word $y_t$ (we assume that the aligns are always correct and align error problem is beyond the focus of this work), 
we expect that a certain portion of ${\bf c}_t$ and the embedding of $y_t$ are semantically equivalent (that is how the information of the source side is transformed to the target side). Therefore, there should be always a certain relation among attention vectors, which are aligned to the same target word. Averaging the attention vectors in different source sentences is expected to highlight the shared portion (\ie corresponds to $y_t$) and dilute the unshared parts (\ie correspond to the contexts of different source sentences).

\begin{table*}[t]
\centering
\begin{tabular}{c|crr c crr c crr}
\multirow{3}{*}{\bf Domain}	&	\multicolumn{3}{c}{\bf Training}	&	&	\multicolumn{3}{c}{\bf Tuning}	&	&	\multicolumn{3}{c}{\bf Test}\\
\cline{2-4}\cline{6-8}\cline{10-12}
				&	\multirow{2}{*}{$|S|$}	&	\multicolumn{2}{c}{$|W|$}	&	&	\multirow{2}{*}{$|S|$}	&	\multicolumn{2}{c}{$|W|$}	&	&	\multirow{2}{*}{$|S|$}	&	\multicolumn{2}{c}{$|W|$}\\
\cline{3-4}\cline{7-8}\cline{11-12}
				&	&	Zh	&	En	&	&	&	Zh	&	En	&	&	&	Zh	&	En\\
\hline

News	&	1.25M	&	27.9M	&	34.5M	&	&	878	&	22.6K	&	23.7K		&	&	6.8K	&	174.1K	&	186.9K\\
Subtitle	&	2.15M	&	12.1M	&	16.6M	&	&	1.1K	&	6.7K		&	9.2K		&	&	1.2K	&	6.7K		&	9.5K\\
TED		&	0.21M	&	4.1M		&	4.4M		&	&	887	&	21.3K	&	17.5K	&	&	5.5K	&	104.1K	&	92.2K\\
\end{tabular}
\caption{Statistics of sentences ($|S|$) and words ($|W|$). K stands for thousands and M for millions.}
\label{table-statistics}
\end{table*}

\subsection{Training and Inference}

\paragraph{Training}
Two pass strategies have proven useful to ease training difficulty when the model is relatively complicated~\cite{Shen:2016:ACL,Wang:2017:AAAI,Tu:2017:AAAI}. 
Inspired by this, we add the cache to a pre-trained NMT model with fine training of only the new parameters related to the cache.

First, we pre-train a standard NMT model which is able to generate reasonable representations (\ie ${\bf c}_t$ and ${\bf s}_t$) to interact with the cache. Formally, the parameters ${\bm \theta}$ of the standard NMT model are trained to maximize the {\em likelihood} of a set of training examples $\{\left[{\bf x}^n, {\bf y}^n\right]\}_{n=1}^{N}$:
\begin{equation}
\hat{\bm \theta} = \argmax_{\bm \theta}\sum_{n=1}^{N}\log P({\bf y}^n|{\bf x}^n; {\bm \theta})
\label{eqn-standard-training}
\end{equation}
where the probabilities of generating target words are computed by Equation~\ref{eqn-standard-probability}.

Second, we fix the trained parameters $\hat{\bm \theta}$ and only fine train the new parameters $\bm \gamma =\{{\bf U}, {\bf V}, {\bf W}\}$ related to the cache  (\ie Equation~\ref{eqn-context-gate}):
\begin{equation}
\hat{\bm \gamma} = \argmax_{\bm \gamma}\sum_{n=1}^{N}\log P(\mathbf{y}^n|\mathbf{x}^n; \hat{\bm{\theta}}, \bm{\gamma})
\label{eqn-new-training}
\end{equation}
where the probabilities of generating target words are computed by Equation~\ref{eqn-our-probability}, and $\hat{\bm{\theta}}$ is trained parameters via Equation~\ref{eqn-standard-training}.
During training, the representations ${\bf c}_t$ and ${\bf s}_t$ remain the same for a given sentence pair with the fixed NMT parameters, thus the cache can be explicitly trained to learn when to exploit translation history to maximize the overall translation performance.

\paragraph{Inference}
Once a model is trained, we use a beam search to find a translation that approximately maximizes the likelihood, which is the same as standard NMT models. After the beam search procedure is finished, we write to the cache the representations that correspond to the $1$-best output. The reason why we do not use $k$-best outputs or all hypotheses in the beam search is two-fold: (1) we want to improve the translation consistency for the final outputs; and (2) continuous representations suffer less from data sparsity problem, in the scenario of which $k$-best outputs generally works better.
Our premiliary experiments validate our assumption, in which $k$-best outputs or hypotheses does not show improvement over their 1-best counterpart.

\section{Experiment}

\subsection{Setup}

\paragraph{Data}
We carried out Chinese-English translation experiments on multiple domains,
each of which differs from others in topic, genre, style, level of formality, etc. 
\begin{itemize}
\item {\bf News}: The News domain is extracted from LDC corpora.\footnote{LDC2002E18, LDC2003E07, LDC2003E14, LDC2004T07, LDC2004T08 and LDC2005T06.} Most sentences in this corpora are formal articles with syntactic structures such as complicated conjuncted phrases, which make textual translation very difficult. We choose the NIST 2002 (MT02) dataset as tuning set, and the NIST 2003-2008 (MT03-08) datasets as test sets.
\item {\bf Subtitle}: The subtitles are extracted from TV episodes, which are usually simple and short~\cite{Wang:2018:AAAI}. Most of the translations of subtitles do not preserve syntactic structures of their original sentences at all. We randomly select two episodes as the tuning set, and other two episodes as the test set.\footnote{The corpora are available at \url{https://github.com/longyuewangdcu/tvsub}.}
\item{\bf TED}:  The corpora are from the MT track on TED Talks of IWSLT2015~\cite{Cettolo:2012:EAMT}.\footnote{\url{https://wit3.fbk.eu/mt.php?release=2015-01}} 
~\newcite{Koehn:2017:WNMT} point out that NMT systems have a steeper learning curve with respect to the amount of training data, resulting in worse quality in low-resource settings.
The TED talks are difficult to translate for its variety of topics while small-scale training data. We choose the ``dev2010'' dataset as the tuning set, and the combination of  ``tst2010-2013'' datasets as the test set.
\end{itemize}
The statistics of the corpora are listed in Table~\ref{table-statistics}. As seen, the averaged lengths of the source sentences in News, Subtitle, and TED domains are 22.3, 5.6, and 19.5 words, respectively.
We use the case-insensitive 4-gram NIST BLEU score~\cite{Papineni:2002} as evaluation metric, and \emph{sign-test}~\cite{Collins:2005} for statistical significance test.

\paragraph{Models}
The baseline is a re-implemented attention-based NMT system \textsc{RNNSearch}, which incorporates dropout~\cite{Hinton:2012:arXiv} on the output layer and improves the attention model by feeding the lastly generated word.
For training \textsc{RNNSearch}, we limited the source and target vocabularies to the most frequent 30K words in Chinese and English, and employ an unknown replacement post-processing technique~\cite{Jean:2015:ACL,Luong:2015:ACL}.
We trained each model with the sentences of length up to 80 words in the training data. 
We shuffled mini-batches as we proceed and the mini-batch size is 80.
The word embedding dimension is 620 and the hidden layer dimension is 1000.
We trained for 15 epochs using Adadelta~\cite{Zeiler:2012:arXiv}, and selected the model that yields best performances on the validation set.

For our model, we used the same setting as \textsc{RNNSearch} if applicable. 
The parameters of our model that are related to the standard encoder and decoder were initialized by the baseline \textsc{RNNSearch} model and were fixed in the following step.
We further trained the new parameters related to the cache for another 5 epochs. Again, the model that performs best on the tuning set was selected as the final model.

\subsection{Effect of Cache Size}

\begin{table}[t]
\centering
\renewcommand\arraystretch{1.1}
\begin{tabular}{c||ccc|c}
{\bf  Cache}	&	{\bf News}	&	{\bf Subtitle}	&	{\bf TED}    &	{\bf Ave}\\
   \hline
   0		&	38.36	&	27.54		&	8.45	&	24.78\\
   \hline
   25		&	39.34	 &	 28.36	&	{\bf 9.24}	&	{\bf 25.65}\\
   50		&	39.36	 &	28.32		&	9.18	&	25.62\\
   100		&	39.48	 &	28.15		&	9.23	&	25.62\\
   200		&	39.39	&	{\bf 28.39}		 &	8.98	&	25.59\\
   500		&	{\bf 39.56}	&	28.10	 	&	9.07	&	25.58\\
   1000	&	39.37	&	27.90	 	&	8.89	&	25.39\\
\end{tabular}
\caption{Translation performances of different cache sizes on the tuning sets.}
\label{table-memory-sizes}
\end{table}

\begin{table}[t]
\centering
\renewcommand\arraystretch{1.1}
\begin{tabular}{c||ccc|c}
{\bf  Overwrite}	&	{\bf News}	&	{\bf Subtitle}	&	{\bf TED}    &	{\bf Ave}\\
\hline
   \texttimes		&	39.42	&	28.26		&	9.12	&	25.60\\
   \hline
   \checkmark		&	39.34	 &	 28.36	&	9.24	&	{25.65}\\
\end{tabular}
\caption{Effect of the cache overwrite mechanism for slots that correspond to the same target word.}
\label{table-overwrite}
\end{table}

\begin{figure}[t]
\begin{center}
\subfloat[size=25]{
     \includegraphics[width=0.235\textwidth]{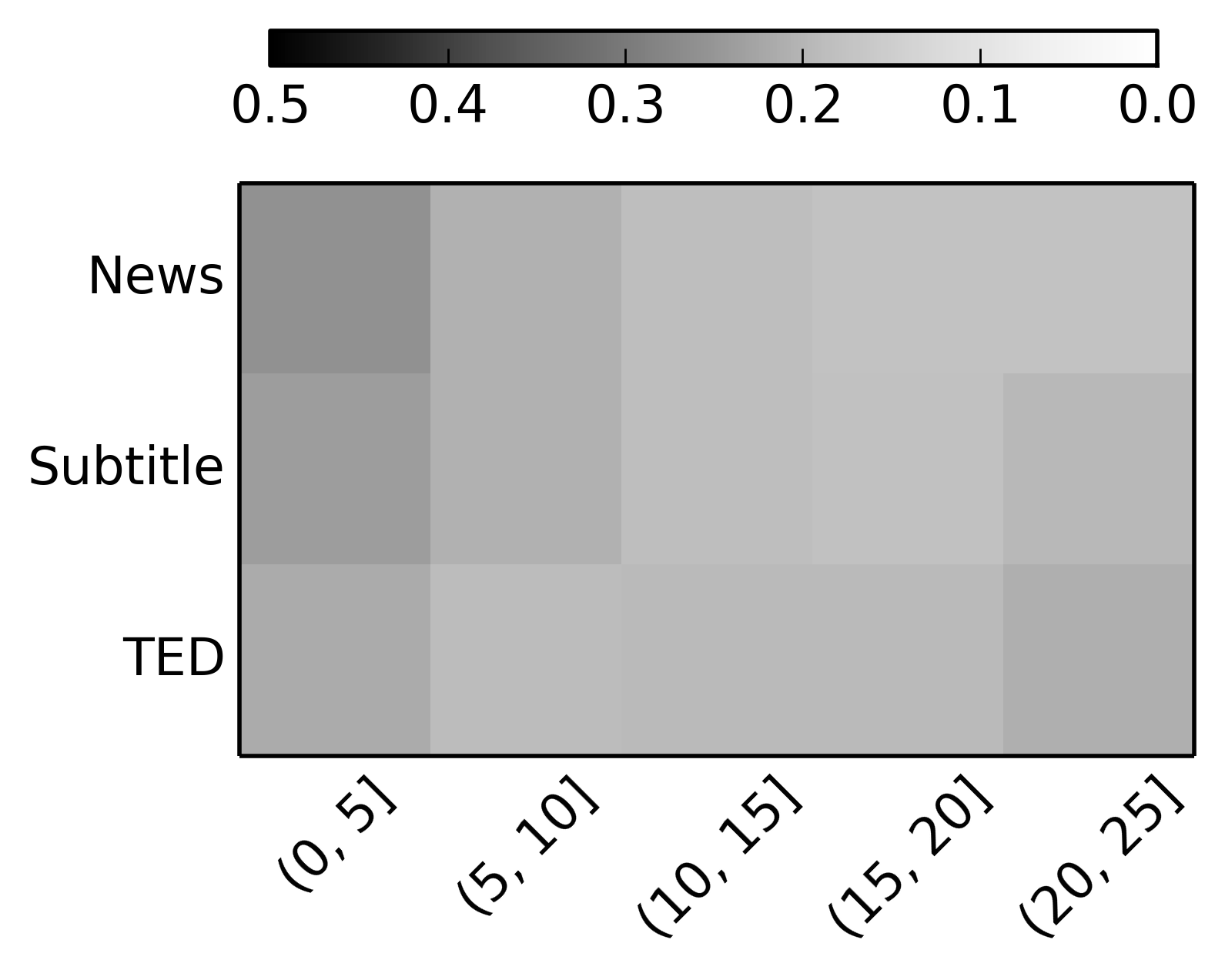}} \hfill
 \subfloat[size=50]{
     \includegraphics[width=0.235\textwidth]{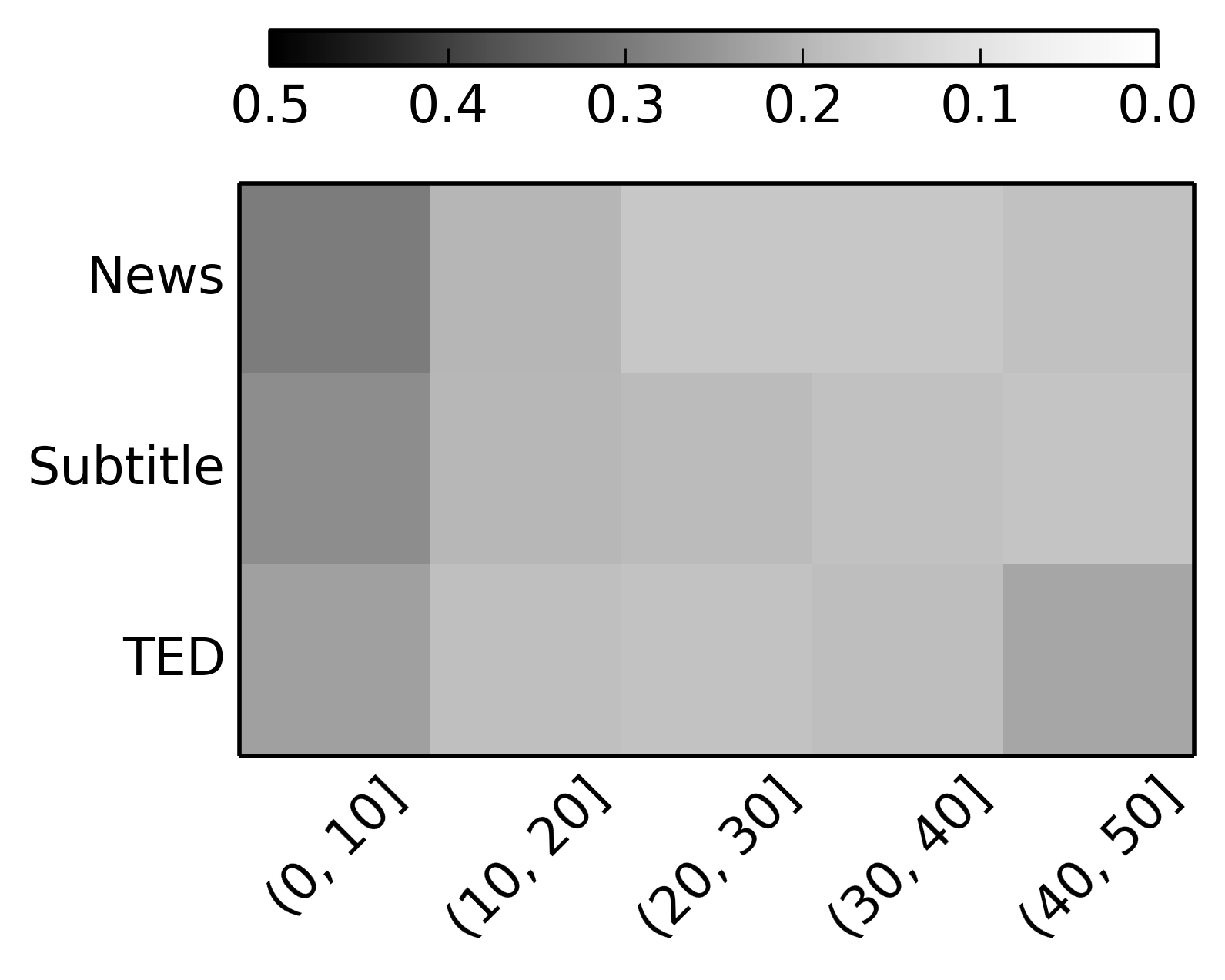}} \\
\subfloat[size=100]{
     \includegraphics[width=0.235\textwidth]{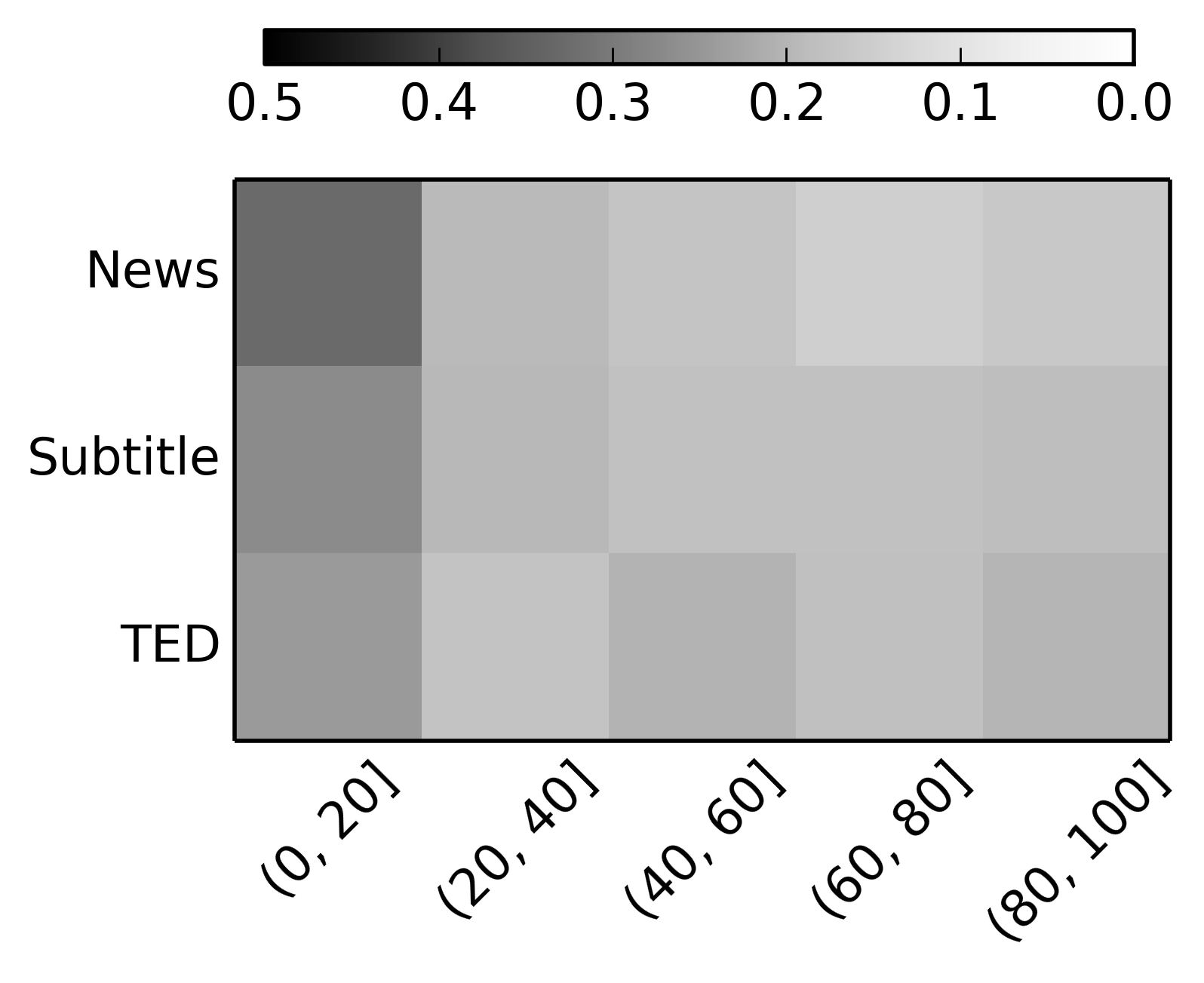}}  \hfill
\subfloat[size=500]{
     \includegraphics[width=0.235\textwidth]{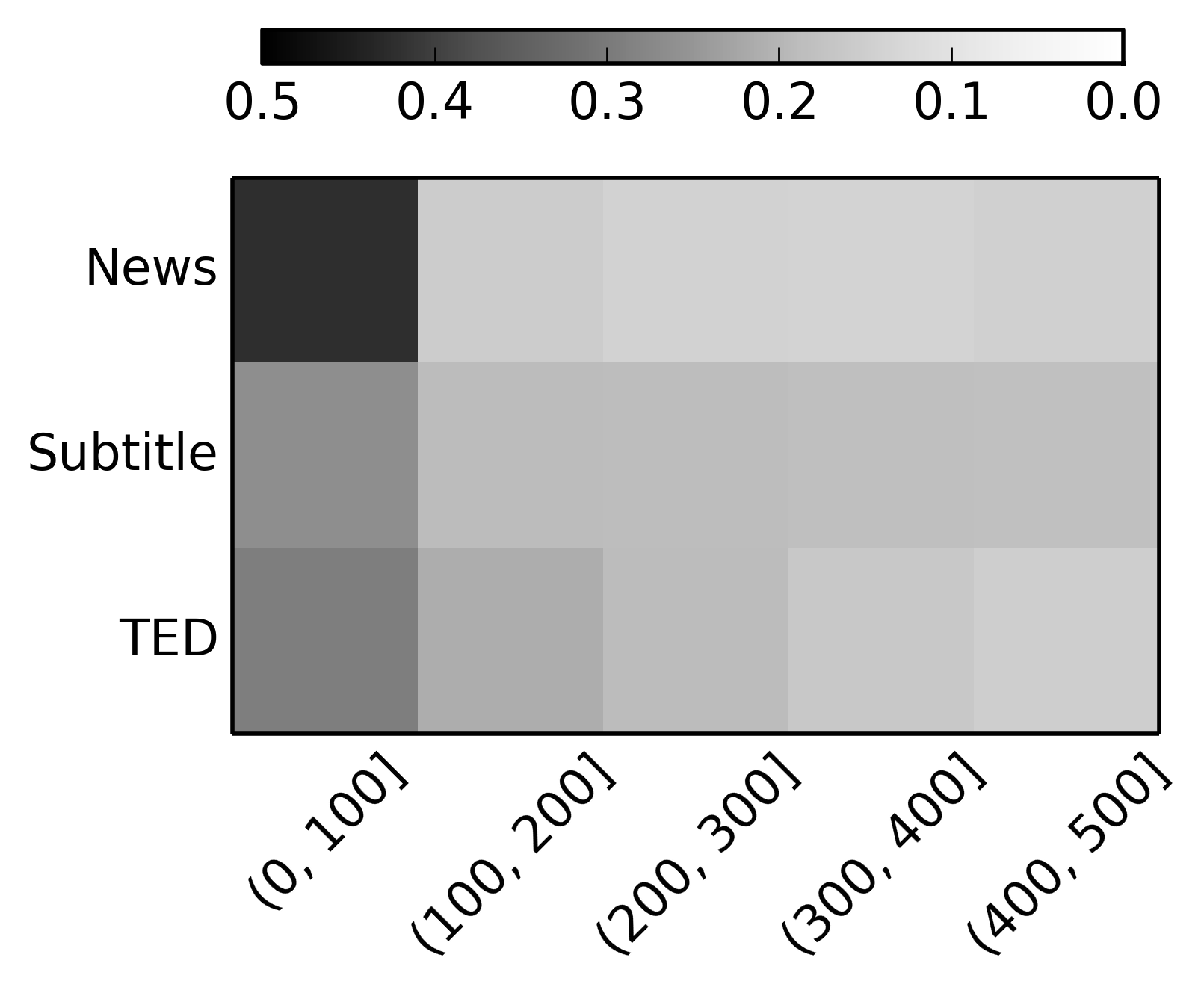}}  \\
\end{center}
\caption{Average cache matching probability distribution on the tune sets,
where the leftmost positions represent the most recent history.} 
\label{figure-distribution}
\end{figure}

Inspired by the recent success of the continuous cache on language modeling~\cite{Grave:2017:ICLR}, we thought it likely that a large cache would benefit from the long-range context, and thus outperforms a small one.
This turned out to be false.
Table~\ref{table-memory-sizes} that lists translation performances of different cache sizes on the tuning set. As seen, small caches (\eg size=25) generally achieve similar performances with larger caches (\eg size=500). 
At the very start, we attributed this to the strength of the cache overwrite mechanism for slots that correspond to the same target word, which implicitly models long-range contexts by combining different context representations of the same target word in the translation history. As shown in Table~\ref{table-overwrite}, the overwrite mechanism contribute little to the good performance of smaller cache.

There are several more possible reasons. First, a larger cache is able to remember longer translation history (\ie cache capacity) while poses difficulty to matching related records in the cache (\ie matching accuracy).
Second, the current caching mechanism fail to model long-range context well, which suggests a better modeling of long-term dependency for future work. 
Finally, neighbouring sentences are more correlated than long-distance sentences, and thus modeling short-range context properly works well~\cite{Daniluk:2017:ICLR}. In the following experiment, we try to validate the last hypothesis by visualizing which positions in the cache are attended most by the proposed model.


\paragraph{Cache Matching Probability Distribution}
Following~\newcite{Daniluk:2017:ICLR}, we plot in Figure~\ref{figure-distribution} the average matching probability the proposed model pays to specific positions in the history. As seen, 
the proposed approach indeed pays more attention to most recent history (\eg the leftmost positions) in all domains. Specifically, the larger the cache, the more attention the model pays to most recent history.

Notably, there are still considerable differences among different domains. For example, the proposed model attends over records further in the past more often in the Subtitle and TED domains than in the News domain.
This maybe because that a talk in the TED testset contains much more words than an article in the News testset (1.9K vs. 0.6K words). Though a scene in the Subtitle testset contains least words (\ie 0.3K words), repetitive words and phrases are observed in neighbouring scenes of the same episode, which is generally related to a specific topic. 
Given that larger caches do not lead to any performance improvement, it seems to be notoriously hard to judge whether long-range contexts are not modelled well, or they are less useful than the short-range contexts. We leave the validation for future work.



For the following experiments, 
the cache size is set to 25 unless otherwise stated.

\begin{table*}[t]
\centering
\renewcommand\arraystretch{1.1}
\begin{tabular}{c||lc|lc|lc|lc}
\multirow{2}{*}{\bf Model}				&	\multicolumn{2}{c|}{\bf News}	&	\multicolumn{2}{c|}{\bf Subtitle}	&	\multicolumn{2}{c|}{\bf TED}	&	\multicolumn{2}{c}{\bf Ave}\\	
   \cline{2-9}
   	&	BLEU	&	$\triangle$		&	BLEU	&	$\triangle$		&	BLEU	&	$\triangle$		&	BLEU	&	$\triangle$\\
   \hline
   \textsc{Base}				&	35.39		&	--		&	32.92		&	--		&	11.69		&	--	&	26.70	&	--\\
   \cite{Wang:2017:EMNLP}	&	{\bf 36.52}$^*$		&	{\bf +3.2\%}	&	33.34		&	+1.0\%	&	12.43$^*$		&	+6.3\%	&	27.43	&	+2.7\%\\
   \cite{Jean:2017:arXiv}		&	36.11$^*$		&	+2.0\%	&	33.00		&	0\%	&	12.46$^*$		&	+6.6\%	&	27.19		&	+1.8\%\\   
   \hline
      \textsc{Ours}				&	36.48$^*$	&	+3.1\%	&	{\bf 34.30}$^*$	&	{\bf +3.9\%}	&	{\bf 12.68}$^*$	&	{\bf +8.5\%}	&	{\bf 27.82}		&	{\bf +4.2\%}\\
\end{tabular}
\caption{Translation qualities on multiple domains. ``*'' indicates statistically significant difference ($p < 0.01$) from``\textsc{Base}'' , and ``$\triangle$'' denotes relative improvement over ``\textsc{Base}''.}
\label{table-in-domain}
\end{table*}

\begin{table}[t]
\centering
\renewcommand\arraystretch{1.1}
\begin{tabular}{c|r|r|r}
\multirow{2}{*}{\bf Model}				&	\multirow{2}{*}{\bf \# Para.}		&	\multicolumn{2}{c}{\bf Speed}\\
 \cline{3-4}
						&			&	{\em Train}	&	{\em Test}\\	
   \hline
   \textsc{Base}				&	84.2M	&	1469.1	&	21.1\\
   \cite{Wang:2017:EMNLP}	&	103.0M	&	300.2		&	20.8\\
   \cite{Jean:2017:arXiv}		&	104.2M 	&	933.8		&	19.4\\
   \hline
   \textsc{Ours}				&	88.2M	&	1163.9	&	21.1\\

\end{tabular}
\caption{Model complexity. ``Speed'' is measured in words/second for both training and testing. We employ a beam search with beam being 10 for testing.}
\label{table-model-complexity}
\end{table}

\subsection{Main Results}

Table~\ref{table-in-domain} shows the translation performances on multiple domains with different textual styles. 
As seen, the proposed approach significantly outperforms the baseline system (\ie \textsc{Base}) in all cases, demonstrating the effectiveness and university of our model. 
We reimplemented the models in~\cite{Wang:2017:EMNLP} and~\cite{Jean:2017:arXiv} on top of the baseline system, which also exploit cross-sentence context in terms of source-side sentences. Both approaches achieve significant improvements in the News and TED domains, while achieve marginal or no improvement in the Subtitle domain.
Comparing with these two approaches, the proposed model consistently outperforms the baseline system in all domains, which confirms the robustness of our approach.
We attribute the superior translation quality of our approach in the Subtitle domain to the exploitation of target-side information, since most of the translations of dialogues in this domain do not preserve syntactic structure of their original sentences at all. They are completely paraphrased in the target language and seem very hard to be improved with only source-side cross-sentence contexts.

Table~\ref{table-model-complexity} shows the model complexity. 
The cache model only introduces 4M additional parameters (\ie related to Equation~\ref{eqn-context-gate}), which is small compared to both the numbers of parameters in the existing model (\ie 84.2M) and newly introduced by~\newcite{Wang:2017:EMNLP} (\ie 18.8M) and~\newcite{Jean:2017:arXiv} (\ie 20M). 
Our model is more efficient in training, which benefit from training cache-related parameters only. To minimize the waste of computation, the other models sort 20 mini-batches by their lengths before parameter updating~\cite{Bahdanau:2015:ICLR}, while our model cannot enjoy the benefit since it depends on the hidden states of preceding sentences.\footnote{To make a fair comparison, which means our model is required to train all the parameters and the other models cannot use mini-batch sorting, the training speeds for the models listed in Table~\ref{table-model-complexity} are 728.8, 159.8, 572.4, and 627.3, respectively.}
Concerning decoding with additional attention models, our approach does not slow down the decoding speed, while~\newcite{Jean:2017:arXiv} decreases decoding speed by 8.1\%. We attribute this to the efficient strategies for cache key matching without any additional parameters.

\subsection{Deep Fusion vs. Shallow Fusion}

Some researchers would expect that storing the words may be a better way to encourage lexical consistency, as done in~\cite{Grave:2017:ICLR}. Following~\newcite{Gu:2017:arXiv}, we call this a {\em Shallow Fusion} at shallow word level, which is in contrast to deep fusion at deep representation level (\ie our approach). We follow~\newcite{Grave:2017:ICLR} to calculate the probability of generating $y_t$ in shallow fusion as
\begin{eqnarray}
P(y_t) &=& (1-\lambda_t)P_{vocab}(y_t) + \lambda_t P_{cache}(y_t) \nonumber \\
P_{cache}(y_t) &=& \mathbbm{1}_{\left\{ y_t = y_i \right\}} P_m ({\bf c}_i | {\bf c}_t) \nonumber
\end{eqnarray}
in which $P_{vocab}(y_t)$ is  the probability of NMT model (Equation~\ref{eqn-standard-probability}) and $P_m ({\bf c}_i | {\bf c}_t)$ is the cache probability (Equation~\ref{eqn-key-matching}). We compute the interpolation weight $\lambda_t$ in the same way as Equation~\ref{eqn-context-gate} except that $\lambda_t$ is a scalar instead of a vector.

\begin{table}[t]
\centering
\renewcommand\arraystretch{1.1}
\begin{tabular}{c|c|c}
{\bf Model}				&	{\bf Tune}		&	{\bf Test}\\
   \hline
   \textsc{Base}				&	38.36	&	35.39\\
   \hline
   Shallow Fusion			&	38.34	&	35.18\\
   Deep Fusion				&	39.34	&	36.48\\
\end{tabular}
\caption{Comparison of shallow fusion (\ie words as cache values) and deep fusion (\ie continuous vectors as cache values)  in the News domain.}
\label{table-shallow-vs-deep}
\end{table}

Table~\ref{table-shallow-vs-deep} lists the results of comparing shallow fusion and deep fusion on the widely evaluated News domain~\cite{Tu:2016:ACL,Li:2017:ACL,Zhou:2017:ACL,WangXing:2017:EMNLP}. As seen, deep fusion significantly outperforms its shallow counterpart, which is consistent with the results in~\cite{Gu:2017:arXiv}. Different from~\newcite{Gu:2017:arXiv}, the shallow fusion does not achieve improvement over the baseline system. One possible reason is that the generated words are less repetitive with those in the translation history than those similar sentences retrieved from the training corpus. Accordingly, storing words in the cache encourages lexical consistency at the cost of introducing noises, while storing continuous vectors is able to improve this problem by doing fusion in a soft way. In addition, the continuous vectors can store useful information beyond a single word, which we will show later.

\begin{CJK}{UTF8}{gbsn}

\begin{figure*}[p]
\centering
\subfloat[We italicize some {\em \color{red} mis-translated} errors and highlight the {\bf \color{blue} correct} ones in bold. Our approach is able to correct the errors with the target-side context retrieved from the cache, as shown below.]{
\renewcommand\arraystretch{1.1}
\begin{tabular}{|c|m{13cm}|}
\hline 
Input & 大家 {\bf 觉得} 这 也是 一次 {\bf 机遇} ， 一次 挑战 。\\\hdashline
Reference & everyone {\bf felt} that this was also an {\bf opportunity} and a challenge .\\\hdashline
\textsc{Base} & we {\em  \color{red} feel} that it is also a {\em \color{red} challenge} and a challenge .\\\hdashline
\textsc{Ours} & everyone {\bf \color{blue} felt} that this was an {\bf \color{blue} opportunity} and a challenge .\\
\hline
Input & 这 确实 是 {\bf 中国队} 不能 “ 善终 ” 的 一 个 原因 。\\\hdashline
Reference & this is indeed a reason why {\bf the chinese team} could not have a `` good ending . ''\\\hdashline
\textsc{Base} & this is indeed the reason why {\em \color{red} china} can not be `` hospice . ''\\\hdashline
\textsc{Ours} & this is indeed a reason why {\bf \color{blue} the chinese team} cannot be `` hospice . ''\\
\hline
\end{tabular}
}\\
\subfloat[Visualization of cache matching matrix, in which the x-axis is the generated words and the y-axis is the cache slots indicated by the corresponding word in translation history. Our approach improves performance by retrieving useful information (\eg verb tense for ``felt'' and phrasal-level patterns for ``the chinese team'', in boxes with red frames) from the cache.]{
     \includegraphics[width=0.33\textwidth]{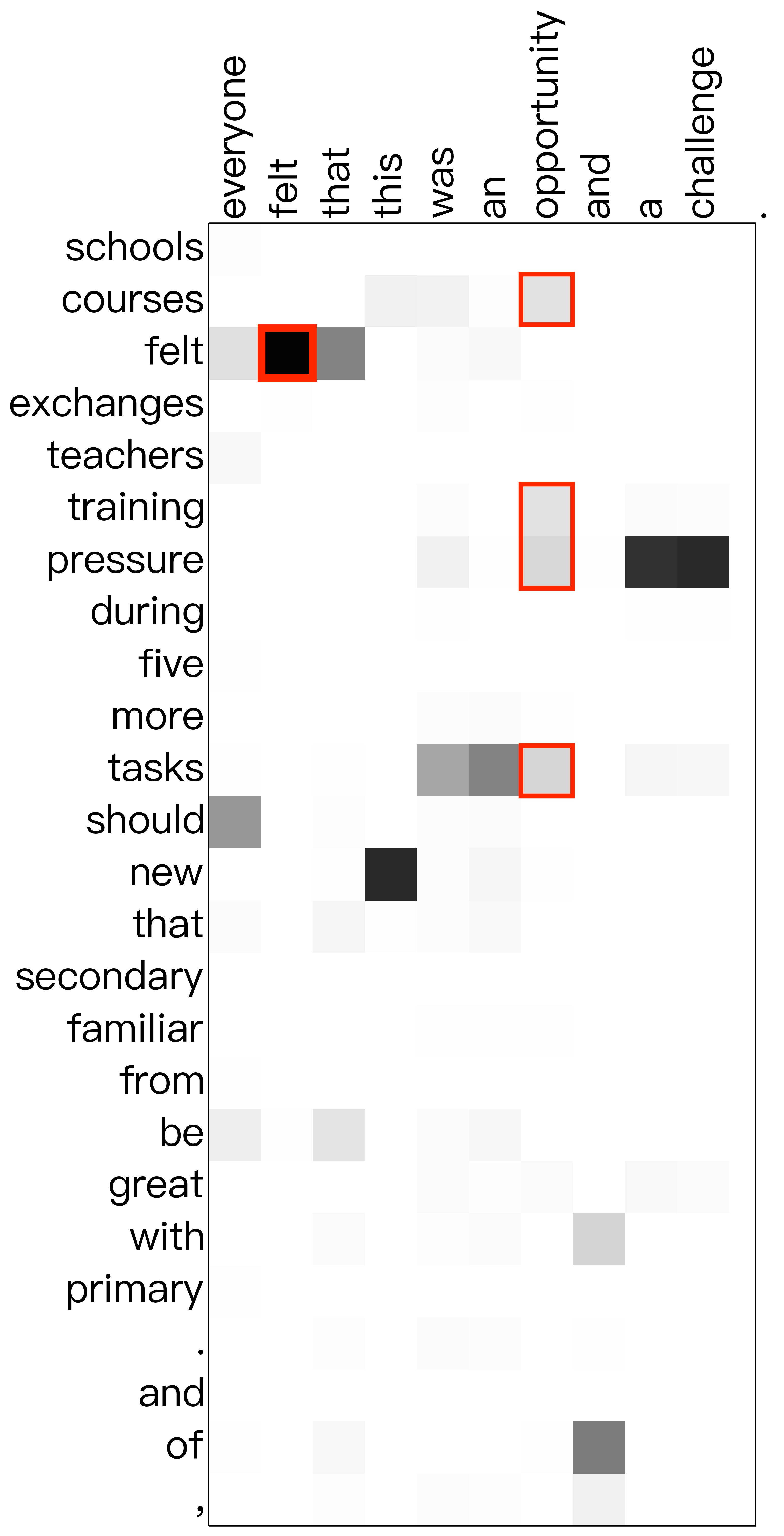}  \hspace{0.15\textwidth}
     \includegraphics[width=0.44\textwidth]{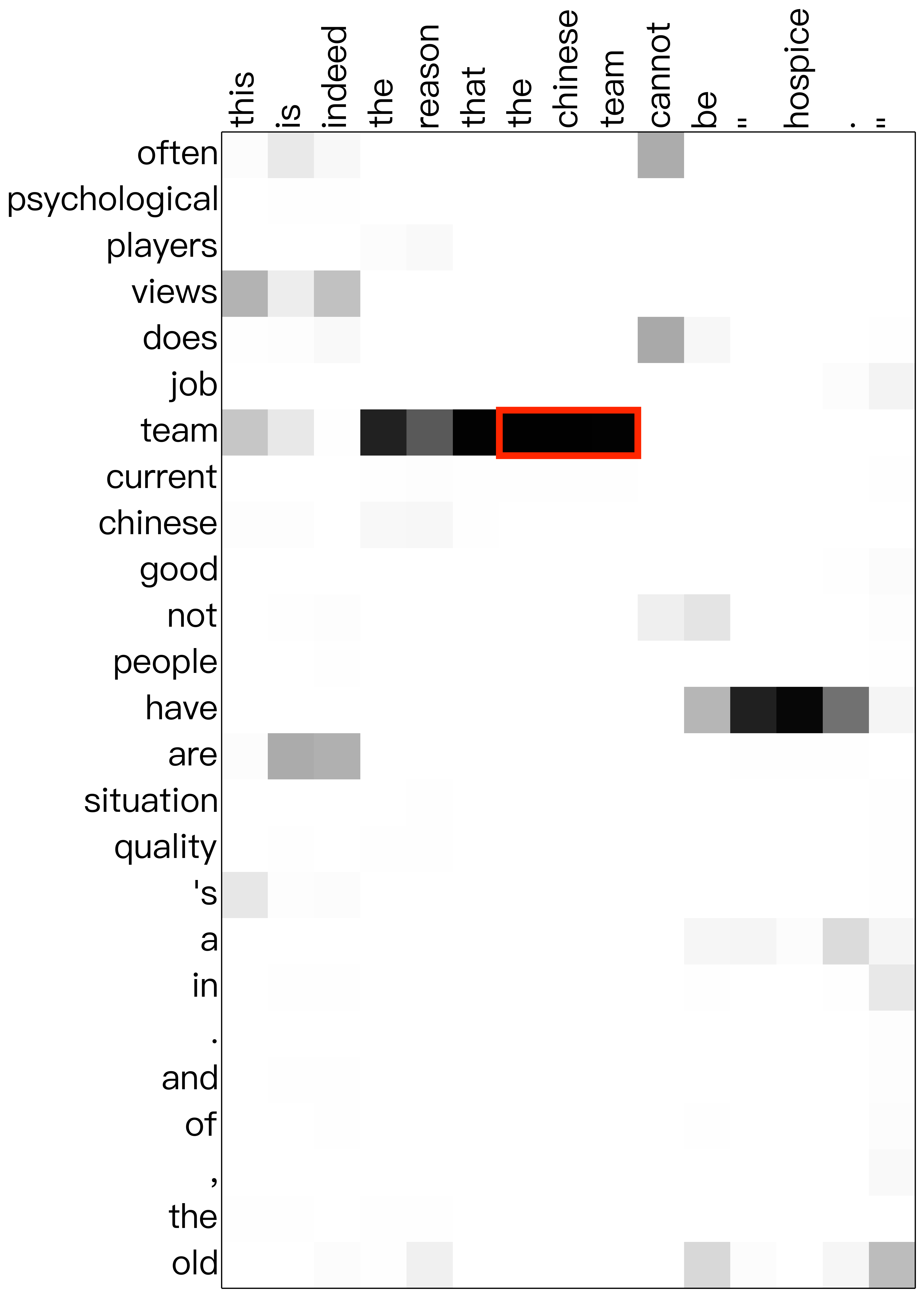}
}\\
\subfloat[Bilingual snippets in translation history that correspond to the slots in boxes with red frames.]{
\renewcommand\arraystretch{1.1}
\begin{tabular}{|c|c|}
\hline 
\dots {\bf \color{blue} 觉得} 新 课程 \dots	&  \dots ， 然后 {\bf \color{blue} 中国队} 将 比分 \dots \\
\dots {\bf \color{blue} felt} that new courses \dots &	\dots ,  {\bf \color{blue} the chinese team} won the \dots \\
\hdashline
\dots 在 听完 了 所有 {\bf \color{blue}  培训} 课程 后  \dots  & \dots {\bf \color{blue} 中国队} 经常 是 在 形势 大好 \dots \\
\dots after listening to all {\bf \color{blue} training} courses \dots &  \dots {\bf \color{blue} the chinese team} often does not have a \dots \\
\hline
\end{tabular}
} 
\caption{Translation examples in which the proposed approach shows its ability to remember translation patterns at different levels of granularity and matching.} 
\label{figure-case}
\end{figure*}

\end{CJK}

\begin{CJK}{UTF8}{gbsn}

\subsection{Translation Patterns Stored in the Cache}

In this experiment, we present analysis to gain insight about what kinds of translation patterns are captured by the cache to potentially improve translation performance, as shown in Figure~\ref{figure-case}.

\paragraph{Tense Consistency}
Consistency is a critical issue in document-level translation, where a repeated term should keep the same translation throughout the whole document \cite{Xiao:2011:MTSummit,Carpuat:2012:WMT}. 
Among all consistency cases, we are interested in the verb tense consistency.
We found our model works well on improving tense consistency. 
For example, the baseline model translated the word ``觉得'' into present tense ``{\em feel}'' in present tense (Figure~\ref{figure-case}(a)), while from the translation history (Table~\ref{table-problems}) we can learn it should be translated into ``felt'' in past tense.
The cache model can improve tense consistency by exploring document-level context. As shown in the left panel of Figure~\ref{figure-case}(b), the proposed model generates the correct word ``{\em felt}'' by attending to the desired slot in the cache. It should be emphasized that our approach is still likely to generate the correct word even without the cache slot ``felt'', since the previously generated word ``{\em everyone}'' already attended to a slot ``{\em should}'', which also contains information of past tense.
The improvement of tense consistency may not lead to a significant increase of BLEU score, but is very important for user experience.

\paragraph{Fuzzily Matched Patterns}
Besides exactly matched lexical patterns (\eg the slot ``felt''), we found that the cache also stores useful ``fuzzy match'' patterns, which can improve translation performance by acting as some kind of ``indicator'' context. 
Take the generation of ``{\em opportunity}'' in the left panel of Figure~\ref{figure-case}(b) as an example, although the attended slots ``{\em courses}'', ``{\em training}'', ``{\em pressure}'', and ``{\em tasks}'' are not matched with ``{\em opportunity}'' at lexical level, they are still helpful for generating the correct word ``{\em opportunity}'' when working together with the attended source vector centering at ``{机遇}''.

\paragraph{Patterns Beyond Word Level}
By visualizing the cache during translation process, we found that the proposed cache is able to remember not only word-level translation patterns, but also phrase-level translation patterns, as shown in the right panel of Figure~\ref{figure-case}(b). The latter is especially encouraging to us, since phrases play an important role in machine translation while it is difficult to integrate them into current NMT models~\cite{Zhou:2017:ACL,WangXing:2017:EMNLP,Huang:2017:arXiv}.
We attribute this to the fact that decoder states, which serve as cache values, stores phrasal information due to the strength of decoder RNN on memorizing short-term history (\eg previous few words).

\end{CJK}

\section{Related Work}

Our research builds on previous work in the field of {\em memory-augmented neural networks}, {\em exploitation of cross-sentence contexts} and {\em cache in NLP}.

\paragraph{Memory-Augmented Neural Networks}
Neural Turing Machines~\cite{Graves:2014:arXiv} and Memory Networks~\cite{Weston:2015:ICLR,Sukhbaatar:2015:NIPS} are early models that augment neural networks with a possibly large external memory.
Our work is based on the Memory Networks, which have proven useful for question answering and document reading tasks~\cite{Weston:2016:ICLR,Hill:2016:ICLR}. Specifically, we use  Key-Value Memory Network~\cite{Miller:2016:EMNLP}, which is a simplified version of Memory Networks with better interpretability and has yielded encouraging results in document reading~\cite{Miller:2016:EMNLP}, question answering~\cite{Pritzel:2017:arXiv} and language modeling~\cite{Tran:2016:NAACL,Grave:2017:ICLR,Daniluk:2017:ICLR}. 
We use the memory to store information specific to the translation history so that this information is available to influence future translations.
Closely related to our approach, ~\newcite{Grave:2017:ICLR} use a continuous cache to improve language modeling by capturing longer history.
We generalize from the original model and adapt it to machine translation: we use the cache to store bilingual information rather than monolingual information, and release the hand-tuned parameters for cache matching.

In the context of neural machine translation,~\newcite{Kaiser:2017:ICLR} use an external key-value memory to remember rare training events in test time, and~\newcite{Gu:2017:arXiv} use a memory to store a set of sentence pairs retrieved from the training corpus given the source sentence. 
This is similar to our approach in exploiting more information than a current source sentence with a key-value memory. Unlike their approaches, ours aims to {\em learning to remember translation history} rather than incorporating arbitrary meta-data, which results in different sources of the auxiliary information (\eg previous translations vs. similar training examples). Accordingly, due to the different availability of target symbols in the two scenarios, different strategies of incorporating the retrieved values from the key-value memory are adopted: hidden state interpolation~\cite{Gulcehre:2016:ACL} performs better in our task while word probability interpolation~\cite{Gu:2016:ACL} works better in~\cite{Gu:2017:arXiv}.

\paragraph{Exploitation of Cross-Sentence Context}
Cross-sentence context, which is generally encoded into a continuous space using a neural network, has a noticeable effect in various deep learning based NLP tasks, such as language modeling~\cite{Ji:2015:ICLR,Wang:2016:ACL}, query suggestion~\cite{Sordoni:2015:ICIKM}, dialogue modeling~\cite{Vinyals:2015:DLW,Serban:2016:AAAI}, and machine translation~\cite{Wang:2017:EMNLP,Jean:2017:arXiv}.

In statistical machine translation, cross-sentence context has proven useful for alleviating inconsistency and ambiguity arising from a single source sentence. 
Wide-range context is firstly exploited to improve statistical machine translation models~\cite{Gong:2011:EMNLP,Xiao:2012:ACL,Hardmeier:2012:EMNLP,Hasler:2014:EACL}. 
Closely related to our approach,~\newcite{Gong:2011:EMNLP} deploy a {\em discrete} cache to store bilingual phrases from the best translation hypotheses of previous sentences. In contrast, we use a {\em continuous} cache to store bilingual representations, which are more suitable for neural machine translation models. 

Concerning neural machine translation,~\newcite{Wang:2017:EMNLP} and~\newcite{Jean:2017:arXiv} are two early attempts to model cross-sentence context.~\newcite{Wang:2017:EMNLP} use a hierarchical RNN to summarize the previous $K$ (\eg $K=3$) source sentences, while~\newcite{Jean:2017:arXiv} use an additional set of an encoder and attention model to encode and select part of the previous source sentence for generating each target word. 
While their approaches only exploit source-side cross-sentence contexts, the proposed approach is able to take advantage of bilingual contexts by directly leveraging continuous vectors to represent translation history.
As shown in Tables~\ref{table-in-domain} and ~\ref{table-model-complexity}, comparing with their approaches, the proposed approach is more robust in improving translation performances across different domains, and is more efficient in both training and testing. 

\paragraph{Cache in NLP}

In NLP community, the concept of ``cache'' is firstly introduced by~\cite{Kuhn:1990:PAMI}, which augments a statistical language model with a cache component and assigns relatively high probabilities to words that occur elsewhere in a given text. The success of the cache language model in improving word prediction rests on capturing of ``burstiness'' of word usage in a local context. It has been shown that caching is by far the most useful technique for perplexity reduction over the standard $n$-gram approach~\cite{Goodman:2001:CSL}, and becomes a standard component in most LM toolkits, such as IRSTLM~\cite{Federico:2008:Interspeech}.
Inspired by the great success of caching on language modeling,~\newcite{Nepveu:2004:EMNLP} propose to use a cache model to adapt language and translation models for SMT systems, and~\newcite{Tiedemann:2010:DANLP} apply an exponentially decaying cache for the domain adaptation task. In this work, we have generalized and adapted from the original discrete cache model, and integrate a ``continuous'' variant into NMT models.

\section{Conclusion}

We propose to augment NMT models with a cache-like memory network, which stores translation history in terms of bilingual hidden representations at decoding steps of previous sentences. 
The cache component is an external key-value memory structure with the keys being attention vectors and values being decoder states collected from translation history. 
At each decoding step, the probability distribution over generated words is updated online depending on the history information retrieved from the cache with a query of the current attention vector. Using simply a dot-product for key matching, this history information is quite cheap to store and can be accessed efficiently.

In our future work, we expect several developments that will shed more light on utilizing long-range contexts, e.g., designing novel architectures, and employing discourse relations instead of directly using decoder states as cache values.


\section*{Acknowledgments}
Yang Liu is supported by the National Key R\&D Program of China (No. 2017YFB0202204) and National Natural Science Foundation of China (No. 61432013, No. 61522204).

\balance
\bibliography{all}
\bibliographystyle{acl2012}

\end{document}